\documentclass[12pt,a4paper]{article}
\usepackage{amsmath, caption, longtable, paralist, amsfonts, amssymb, setspace}
\usepackage[top=1.5in, bottom=1.5in, left=1.1in, right=1.1in]{geometry}
\DeclareCaptionStyle{italic}[justification=centering]{labelfont={bf}, textfont={it}, labelsep=colon}
\usepackage{psfrag, epsf}
\usepackage[dvipsnames]{xcolor}
\usepackage{enumerate}
\usepackage{url} 
\usepackage{caption}
\usepackage{tabularx}
\usepackage{subcaption}
\usepackage{bbm, dsfont, mathpazo, bm}
\usepackage{comment}
\usepackage{amsthm}
\usepackage{float}
\usepackage{enumitem}
\usepackage{booktabs, microtype, rotating} 
\usepackage[pdftex,colorlinks=true]{hyperref}
\definecolor{darkblue}{rgb}{0,0,.6}
\hypersetup{citecolor=darkblue,linkcolor=darkblue,urlcolor=darkblue}
\allowdisplaybreaks

\newcommand{\blind}{0}

\usepackage{xcolor,colortbl}

\definecolor{Gray}{gray}{0.85}
\definecolor{LightCyan}{rgb}{0.88,1,1}

\newcolumntype{a}{>{\columncolor{Gray}}c}
\newcolumntype{b}{>{\columncolor{white}}c}

\usepackage{mathtools}
\usepackage{graphicx}
\usepackage{multirow}

\usepackage{algpascal}
\usepackage{amsmath, amssymb, amsthm}
\usepackage{textcomp}
\usepackage[export]{adjustbox}

\usepackage{tikz}

\let\proglang=\textsf
\usepackage{smartdiagram}
\usepackage[utf8]{inputenc}
\usetikzlibrary{arrows,positioning}
\usepackage{makecell} 
\setcellgapes{5pt}

\graphicspath{{plots/}}
\newsavebox\CBox

\usepackage{orcidlink}
\date{}
\AtBeginDocument{}
\usepackage{cleveref}
\usepackage{algorithm}
\usepackage{algpseudocode}
\usepackage[authoryear,round]{natbib}
\setcitestyle{aysep={}}
\usepackage{notoccite}
\begin{document}
	
\def\spacingset#1{\renewcommand{\baselinestretch}{#1}\small\normalsize} \spacingset{1}

\if0\blind
{
	\title{\bf Amortized Neural Clustering of Time Series based on Statistical Features}
	\author{\'{A}ngel L\'{o}pez-Oriona\orcidlink{0000-0003-1456-7342}\thanks{Postal address: CEMSE Division, Statistics Program, King Abdullah University of Science and Technology (KAUST), Thuwal 23955-6900, Saudi Arabia. E-mail: angel.lopezoriona@kaust.edu.sa} \   and Ying Sun\orcidlink{0000-0001-6703-4270}  \hspace{.2cm}\\  
		Statistics Program\\ 
		King Abdullah University of Science and Technology (KAUST)\\ Thuwal, Saudi Arabia \\ 
	}
	\maketitle
} \fi

\if1\blind
{
	\title{\bf Amortized Neural Clustering of Time Series based on Statistical Features}
} \fi

\bigskip
		
\begin{abstract}
This paper introduces an algorithm-agnostic approach to feature-based time series clustering via amortized neural inference. By training neural networks to approximate the optimal partitioning rule from simulated data, the proposed framework reduces reliance on conventional clustering methods, such as $K$-means, $K$-medoids, or hierarchical clustering, and their associated objective functions and heuristics. Leveraging statistical features, such as autocorrelations and quantile autocorrelations, the approach learns a data-driven affinity structure from which clustering partitions can be recovered, without requiring explicit prior specification of cluster shapes or structures. In addition, one version of the method can automatically determine the number of clusters, avoiding ad-hoc selection procedures. Comprehensive empirical studies show that the proposed framework achieves competitive or superior clustering accuracy relative to traditional methods, even in challenging scenarios where competing techniques are provided with the true number of clusters. An application to financial time series of stock returns illustrates its practical utility. By reducing the need for algorithm selection and calibration, the proposed framework opens new possibilities for automated, adaptive, and data-driven clustering of temporal data across scientific and industrial domains.
		
		\vspace{.1in}
		
	\end{abstract}
	
	\doublespacing

    \section{Introduction}\label{sectionintroduction}
 
	The vast increase in time series data available today, originating from fields as diverse as finance, healthcare, and environmental science, has created a strong demand for methods that can efficiently analyze and make sense of large collections of sequences. Time series clustering is a widely used unsupervised learning technique designed to tackle this challenge by grouping similar series together, usually without the need to model each one individually, a process that can be both computationally expensive and unnecessary for many applications. Instead, clustering aims to reveal common patterns that summarize the behavior of the data, supporting tasks such as pattern recognition, anomaly detection, and classification. Over the years, research in time series clustering has grown substantially, evolving from classical approaches based on distance measures to more recent developments leveraging deep learning. Several comprehensive reviews provide overviews of this evolution, including \cite{fu2011review, aghabozorgi2015time, maharaj2019time}, and the latest survey by \cite{Paparrizos_et_al_2024}, which covers both traditional methods and modern neural network techniques.
	
	Feature-based methods are among the most widely used approaches for time series clustering. These techniques typically consist of two key stages: (i) extracting a feature vector for each time series, which captures statistical properties that often reflect the underlying temporal dependencies, and (ii) applying a conventional clustering algorithm to the set of feature vectors. Feature-based clustering of time series has been thoroughly explored, with methods leveraging model-based features \citep{piccolo1990distance, d2013autoregressive}, wavelet-based features \citep{ann2010wavelet, d2012wavelets}, quantile-based features \citep{lafuente2016clustering, lafuente2020robust, lopez2021quantile, lopez2021spatial}, and features based on extremes \citep{d2017fuzzy}, among others, each offering distinct advantages in capturing different aspects of time series dynamics.
	Despite its success, classical feature-based clustering of time series has several important limitations, which we outline below:

	\begin{itemize}
		\item \textbf{Feature selection}. The analyst must determine a set of features with high discriminative power (ideally, optimal) to differentiate between the underlying structures, which is a challenging task. This is typically achieved through domain knowledge combined with exploratory analysis.
		\item \textbf{Algorithm selection}. Once a given set of features has been selected, the analyst must decide which algorithm to use for clustering in the feature space. For example, in the case of hard (also referred to as crisp) clustering, where each time series is assigned to only one group with no overlap between clusters, common choices include $K$-means, $K$-medoids, hierarchical clustering, etc., all of which have several variants. Each method has its own advantages and disadvantages \citep{madhulatha2011comparison, rodriguez2019clustering}, making it difficult to determine in advance which will be the best choice for the set of time series under analysis. Moreover, each method typically aims to optimize a specific objective (e.g., within-cluster compactness in the case of $K$-means and $K$-medoids), which may not be suitable for the particular characteristics of the analyzed data.
        \item \textbf{Selection of the number of clusters}. In practice, the true number of groups is almost always unknown, yet many popular algorithms (such as $K$-means and $K$-medoids) require this parameter as input. Although criteria based on internal clustering quality indices \citep{liu2010understanding} are commonly used, determining the optimal number of clusters remains an essential open problem in the clustering literature \citep{mirkin2011choosing, schubert2023stop}.
		\item \textbf{Optimization limitations}. Once the feature dataset is provided as input to the chosen clustering algorithm (even if the analyst has selected the optimal features, the optimal algorithm, and the true value of the number of groups), there is often no guarantee that the procedure will return the ``best'' clustering partition given the observed data. For instance, it is well-known that traditional $K$-means and $K$-medoids algorithms do not ensure convergence to a global minimum. Several heuristics are often used to mitigate this issue \citep{fong2014towards, harris2022extensive}, such as running the corresponding algorithm with multiple random initializations and picking the solution associated with the lowest value of the objective function. While helpful, these heuristics often fail in complex scenarios. 
	\end{itemize}
	
	While the first point above constitutes a major problem spanning extensive literature, not limited to the temporal setting~\citep{alelyani2013feature, kuzudisli2023review}, addressing it proves complex, as feature selection often depends on domain knowledge and the specific clustering goals (different features can yield distinct interpretations of the clustering solution). Thus, this article focuses on the remaining challenges. Specifically, our objective is to develop a mechanism that: (i) performs accurate, automated time series clustering while reducing reliance on predefined clustering objectives, (ii) can adapt to varying numbers of clusters in a data-driven manner, at least in scenarios with simple to moderate complexity, and (iii) mitigates typical optimization drawbacks of traditional clustering procedures. To this end, we adopt the amortized inference framework (see Section~\ref{sectionimprovingtsc}).

	The remainder of this paper is organized as follows. In Section \ref{sectionimprovingtsc}, we explore how the amortized inference framework can be adapted to the particular case of feature-based time series clustering. Section \ref{sectionsimulations} presents some empirical results, while Section \ref{sectionapplication} shows one application to real-world financial time series of stock returns. Conclusions are given in Section \ref{sectiondiscussion}. 
	
	\section{Improving time series clustering via amortized inference}\label{sectionimprovingtsc}
	
	This section reviews the fundamental concepts of simulation-based and amortized inference, and then explains how these frameworks can be adapted for feature-based time series clustering.
	
	\subsection{Background on simulation-based and amortized inference}\label{reviewai}
	
   Simulation-based inference encompasses a broad class of techniques that enable statistical inference by leveraging repeated simulations from complex, often intractable, data generating models. When explicit likelihood functions are unavailable or computationally prohibitive, these methods approximate inferential quantities by comparing summary statistics of simulated data to those of observed data, circumventing the need for closed-form likelihood evaluations \citep{papamakarios2019sequential, cranmer2020frontier}. Classical approaches such as approximate Bayesian computation and Bayesian synthetic likelihood rely on kernel-based or parametric approximations of the likelihood constructed from simulations \citep{beaumont2002approximate, price2018bayesian}. These techniques have found widespread application in domains like biology, ecology, and particle physics, where models are typically stochastic and analytically intractable \citep{beaumont2010approximate, gonccalves2020training}. However, traditional simulation-based methods often require a large number of simulations, which can be computationally expensive or infeasible, particularly for high-dimensional data or costly simulators \citep{cranmer2020frontier}.

   To overcome the limitations of traditional simulation-based inference methods, recent advances have introduced neural simulation-based inference techniques that leverage deep learning to efficiently capture the relationship between observed data and model parameters. Among these, amortized inference has emerged as a powerful framework. It trains a single parameterized function, typically a deep neural network, to approximate the inference task across a broad range of datasets, thereby distributing the computational cost over multiple inference queries~\citep{kingma2013auto,zammit2025neural}. The term ``amortized'' embodies spreading computational effort: instead of expensive inference from scratch for every new dataset, most computation occurs upfront during training, enabling rapid inference on new observations. This reformulates inference by learning a function that maps data to solutions like posteriors, generalizing across datasets and eliminating repeated costly procedures (unlike MCMC or variational inference's iterations per dataset) trading upfront costs for fast, scalable inference. This advantage makes amortized inference particularly well-suited not only for large-scale and streaming data applications but also for complex tasks like clustering, where fast and adaptive inference of data structure is critical.
   
   Contemporary amortized inference approaches increasingly exploit advances in neural architectures, optimization algorithms, and computing hardware to approximate complex inference tasks. These methods aim to learn flexible mappings from observed data to inferential quantities, benefiting from modern machine learning tools. A recent comprehensive review by Zammit-Mangion et al. \citep{zammit2025neural} synthesizes progress in this area, covering developments in point estimation, approximate likelihood modeling, and summary-statistic learning. The review also contrasts neural amortized inference with traditional sampling-based techniques, highlighting improvements in computational efficiency and scalability enabled by modern machine learning frameworks and hardware acceleration.
   
   While originally motivated by Bayesian posterior estimation, both simulation-based and amortized inference have evolved into general-purpose tools for probabilistic modeling. Their flexibility and computational advantages make them valuable for a wide range of applications, facilitating efficient and automated inference beyond the capabilities of classical paradigms \citep{papamakarios2019sequential, zammit2025neural}.
	
	\subsection{Adapting the amortized inference framework for time series clustering}\label{aitsc}
	
    Let $\mathbb{X} = \bigl\{ \bm{X}_1, \bm{X}_2, \ldots, \bm{X}_n \bigr\} \in \mathcal{X}$ be a set of $n$ time series, where $\mathcal{X}$ denotes the collection of all sets of $n$ time series. Hereafter, we assume for simplicity that $\mathbb{X}$ contains real-valued, univariate time series, each of common length $T$, although the proposed methodology applies to time series of any type and varying lengths. Given a number of clusters \(K\le n\), feature-based clustering is often performed by first extracting the set of feature vectors \(\mathbb{F} = \Big\{\bm \Psi_1, \bm \Psi_2, \ldots, \bm \Psi_n \Big\} \in \mathcal{F}\), with each $\bm \Psi_i \in \mathbb{R}^d$ being a $d$-dimensional vector associated with the $i$th time series, $i=1,\ldots,n$, and then applying a classical clustering algorithm to the set \(\mathbb{F}\), frequently using the number of groups \(K\) as input. Note that the feature vectors in \(\mathbb{F}\) can contain any statistical information from the time series (e.g., mean values, dependence measures such as autocorrelations, model-based quantities, etc.) that are relevant for constructing the clustering structure. Typically, the analyst selects the features using domain knowledge and some preprocessing analyses.
   
   Implicitly, the above approach is based on determining a decision rule \(\bm \delta(\cdot)\) allowing to obtain the clustering partition. Formally, the decision rule can be seen as a function
   
   \begin{equation*}
   	\bm \delta : \mathcal{X} \to \mathcal{C}_K,
   \end{equation*}
   
   \noindent where \(\mathcal{C}_K\) is the set of all possible partitions of the \(n\) time series into \(K\) groups. For instance, when performing feature-based clustering using \(K\)-means, the underlying decision rule we target is given by \(\bm \delta_{\text{$K$-means}} : \mathcal{X} \to \mathcal{C}_K\), such that, for a given collection of time series $\mathbb{X}$, \(\bm \delta_{\text{$K$-means}}(\mathbb{X})\) defines the clustering partition minimizing the objective function of the corresponding standard \(K\)-means problem based on the set $\mathbb{F}$. Note that, in practice, the above mechanism involves solving a separate clustering problem for each collection \(\mathbb{X}\). That is, each time a set of \(n\) time series is analyzed, a different \(K\)-means problem must be solved using standard iterative methods. Moreover, for each clustering problem addressed, there is no guarantee of reaching a global optimum, that is, there is no guarantee of finding $\bm \delta_{\text{$K$-means}}$. A similar reasoning applies to alternative clustering methods, such as \(K\)-medoids or hierarchical clustering.
   
  In the previous context, one often assumes the existence of a collection of \(K\) underlying processes, $\mathbb{G} = \{\bm{G}_1, \bm{G}_2, \ldots, \bm{G}_K\}$, generating the series in \(\mathbb{X}\), for example, autoregressive (AR) processes of a given order. Throughout this paper, we will assume that the features considered to construct the set \(\mathbb{F}\) are rich enough to allow for proper discrimination between the \( K \) generating processes for sufficiently large values of the series length; for instance, estimated autocorrelations at lag 1 if all processes in \(\mathbb{G}\) are causal, Gaussian AR(1). The main goal of clustering under the above circumstances should be obtaining the rule allowing to recover the true clustering structure for any set of time series, which we refer to as the optimal decision rule \(\bm{\delta}^*\). This can be expressed as the function \(\bm{\delta}^* : \mathcal{X} \to \mathcal{C}_K\) such that
  \begin{equation*}
  	\bm{\delta}^*(\mathbb{X}) = \{\pi_1, \pi_2, \ldots, \pi_n\},
  \end{equation*}
  where, for \( i = 1, \ldots, n \), \(\pi_i = k\) if the generating process of the series \(\bm{X}_i\) is \(\bm{G}_k\). However, in most practical scenarios, the optimal decision rule~$\bm{\delta}^*$ is not available in closed form, as it would require complete knowledge of the underlying data generating processes and induced clustering structure, information that is rarely accessible in real applications. Therefore, it becomes necessary to approximate~$\bm{\delta}^*$ using flexible function approximators; recently, amortized inference via neural networks provides a powerful framework for learning such mappings directly from data (see Section \ref{reviewai}), enabling efficient and accurate clustering of new time series collections. In this sense, the proposed framework can be interpreted as learning an approximation to the optimal clustering rule itself, rather than solving a sequence of optimization problems defined by a fixed objective function. 
  
  The above framework can be generalized to the case where the number of time series ($n$) and the number of clusters ($K$) are not fixed. In this scenario, the optimal decision rule would produce the true clustering partition for a set containing an arbitrary number of time series. Under this general setting, the generating mechanism we consider to approximate the optimal decision rule is as follows, where $p(\cdot)$ denotes a generic distribution:
  \begin{equation*}
  	\begin{aligned}
  		& n \sim p(n), \\
  		& \bm \alpha_1 \sim p(\bm \alpha_1), \\
  		& \bm \alpha_2 \sim p(\bm \alpha_2), \\
  		& \pi_1,\, \ldots,\, \pi_n \sim p(\pi_1, \ldots, \pi_n \mid \bm \alpha_1), \\
  		& \bm{\theta}_1,\, \ldots,\, \bm{\theta}_K \mid \pi_1, \ldots, \pi_n \sim p(\bm{\theta}_1,\, \ldots,\, \bm{\theta}_K \mid \bm \alpha_2), \\
  		& \bm{X}_i \sim \bm{G}_{\pi_i} \mid \bm{\theta}_{\pi_i}, \, \, \, i = 1,\, \ldots,\, n,
  	\end{aligned}
  \end{equation*}

  \noindent where $\bm \alpha_1$ and $\bm \alpha_2$ are vectors of hyperparameters and $\bm{\theta}_k$ is a parameter vector determining the form of the $k$th generating process $\bm{G}_k$ (e.g., $\bm{\theta}_k$ includes the autoregressive coefficient and the variance of the error in the case of a zero-mean Gaussian AR process of order 1). Note that the number of clusters, $K$, is, in general, a random variable indicating the number of different values among the generated $\pi_1,\, \ldots,\, \pi_n$.
  
  In practice, one could simulate a large number of time series collections as indicated above and consider the corresponding set of features and the true clustering structure for each collection. Such data can be used to train a neural network to approximate the optimal decision rule. Once the neural network is trained, the clustering solution for a new set of time series can be easily obtained with a single forward pass once the new set of features is calculated. Clearly, the success of such an approach hinges on the quality of the generating scheme and the learning ability of the network. In other words, the generating mechanism must capture the essential characteristics of the clustering problem that the analyst is interested in, while the network architecture must be suitable.
	
\section{Empirical study}\label{sectionsimulations}
	
In this section, we evaluate the performance of the proposed amortized neural clustering method through a comprehensive simulation study, considering scenarios with both fixed and variable numbers of time series and clusters. First, we analyze clustering based on classical autocorrelations, followed by an examination of clustering using quantile autocorrelations. Finally, we assess the method's performance under misspecification of the cluster-defining models.
	
\subsection{Clustering based on autocorrelations}\label{subsectionsimulations1}
    
To generate the synthetic datasets used for the empirical study, we first consider AR(\(p\)) processes as the underlying time series models. A univariate AR(\(p\)) process \(\{X_t, t \in \mathbb{Z}\}\) is defined by the recursion  
    \[
    X_t = \sum_{i=1}^p \phi_i X_{t-i} + \varepsilon_t,
    \]
    where \(\bm \phi=(\phi_1, \phi_2, \ldots, \phi_p)\) are the autoregressive coefficients, and \(\{\varepsilon_t, t \in \mathbb{Z}\}\) is a white noise process with zero mean and variance \(\sigma^2\). The process is stationary and causal if the roots of the characteristic polynomial $\Phi(z) = 1 - \sum_{i=1}^p \phi_i z^i$, $z \in \mathbb{C}$, lie outside the unit circle.
    
    In this experiment, we focus on AR(3) processes as the generating mechanism for each cluster. In each trial, we fix the number of clusters, \(K\), and the number of time series, \(n\). To create cluster-specific generative models, for cluster \(k\), \(k = 1, \ldots, K\), we randomly sample the corresponding AR(3) coefficients, \(\bm{\phi}^k = (\phi_1^k, \phi_2^k, \phi_3^k)\), such that the causality condition is satisfied. To do that, we focus on the corresponding partial autocorrelations \(\bm{\kappa}^k = (\kappa_1^k, \kappa_2^k, \kappa_3^k)\). Additionally, the corresponding error variance \(\sigma_k^2\) is independently sampled from a certain distribution. Specifically, the generating mechanism, which we will refer to as Scenario 1, is as follows:
\begin{align}\label{simm1}
    		& n, K \text{ fixed}, \notag \\
    		& \boldsymbol{\alpha}_1 \sim \text{Dirichlet}(\mathbf{1}_K), \notag \\
    		& \pi_i \mid \boldsymbol{\alpha}_1 \sim \text{Categorical}(\boldsymbol{\alpha}_1), \quad i=1, \ldots, n, \notag \\
    		& \kappa_j^k \sim U[-1,1], \quad j=1,2,3, \quad k=1,\ldots,K, \\
    		& \sigma_k^2 \sim U[0.1, 2], \quad k=1, \ldots, K, \notag \\
    		& \text{Obtain AR coefficients } \boldsymbol{\phi}^k \text{ from } \boldsymbol{\kappa}^k \text{ via Durbin-Levinson recursion}, \quad k=1,\ldots,K, \notag \\
    		& \boldsymbol{X}_i \mid \pi_i, \boldsymbol{\phi}^{\pi_i}, \sigma_{\pi_i}^2 \sim \text{Gaussian AR}(3) \text{ with coefficients } \boldsymbol{\phi}^{\pi_i} \text{ and variance } \sigma_{\pi_i}^2, \quad i=1, \ldots, n, \notag
    	\end{align}
    
    \noindent where $\bm 1_K$ denotes a vector with $K$ ones. Note that the cluster proportions \(\boldsymbol{\alpha}_1\) are drawn from a symmetric Dirichlet distribution, inducing moderately unequal cluster sizes that reflect a realistic but balanced diversity among the clusters. The partial autocorrelation coefficients \(\kappa_j^k\) are sampled uniformly from \([-1,1]\), ensuring a broad coverage of the stationary parameter space for each \( \text{AR}(3) \) process. The corresponding autoregressive coefficients \(\boldsymbol{\phi}^k\) are obtained via the Durbin-Levinson algorithm \citep{durbin1960fitting}, guaranteeing stationarity. Noise variances \(\sigma_k^2\) are drawn from a uniform distribution over \([0.1, 2]\), allowing for a realistic range of innovation noise across clusters. Scenario 1 enables the simulation of diverse but well-structured time series data suitable for comprehensive clustering evaluation.
    
    To train a neural network that approximates the optimal decision rule for the setting described above, we proceed as follows. Given fixed values of $n$ and $K$, we generate a large number ($N$) of time series collections using the simulation mechanism outlined in Equation~\eqref{simm1}, with all series having a common length~$T$. Each collection is associated with ground truth cluster memberships $\pi_1, \ldots, \pi_n$. For each time series, feature extraction is performed by computing the estimated autocorrelations for the first three lags, resulting in a set of $N$ feature datasets. Specifically, for the observations $X_{i,1}, \ldots, X_{i,T}$ of time series $\bm{X}_i$, the estimated autocorrelation at lag $l$ is given by $\widehat{\rho}_i(l)
=\frac{\frac{1}{T}\sum_{t=1}^{T-l} (X_{i,t+l}-\widehat{\mu}_i)(X_{i,t}-\widehat{\mu}_i)}
{\frac{1}{T}\sum_{t=1}^{T}(X_{i,t}-\widehat{\mu}_i)^2}$, where $\widehat{\mu}_i=\frac{1}{T}\sum_{t=1}^{T}X_{i,t}$. The quantity $\widehat{\rho}_i(l)$ serves as a natural estimator of the theoretical autocorrelation 
$\rho_i(l) = \mathrm{Corr}(X_t^{(i)}, X_{t+l}^{(i)})$, where $\{X_t^{(i)}, t \in \mathbb{Z}\} = \bm{G}_{\pi_i}$ denotes the underlying process of the $i$th time series. The resulting feature vector for the $i$th series, $\bm{\Psi}_i$, is then constructed from these estimated autocorrelations. The collection of feature datasets, together with the corresponding true partitions, is subsequently used to train the neural network.

    \subsubsection{Network architecture and predictions}\label{subsubsectionna}
    
    Our approach employs a neural network that predicts the probability that two time series belong to the same cluster, leveraging only extracted statistical features. For each simulated dataset, the network receives the corresponding collection of feature vectors and considers all unordered pairs of time series; for each pair, each series vector is independently passed through a shared embedding network, and their embeddings are symmetrically aggregated by summation to form a permutation-invariant representation. This pairwise embedding is then passed through a multi-layer perceptron (MLP) with a single hidden layer comprising 256 units and a ReLU activation function. The final output layer uses a sigmoid activation to produce a probability score for cluster co-membership. Training is performed using a binary cross-entropy loss, with targets reflecting the true pairwise partition information from the generated data (so the network learns to approximate the optimal co-assignment rule). The network is trained over the $N$ simulated datasets, ensuring exposure to diverse partition structures. Optimization utilizes Adam with a learning rate of 0.001. This pairwise learning setup, grounded in Deep Sets framework \citep{Zaheer2017DeepS}, ensures permutation invariance and enables flexible, algorithm-agnostic clustering by allowing the network to learn the underlying partition structure of the data. Similar pairwise mechanisms have proven highly effective in prior works on deep clustering and meta-learning for unsupervised grouping~\citep{lu2007semi, hsu2015neural, liu2021cluster, sadeghi2024deep}.
    
    After training the neural network, inference on a new set of time series proceeds as follows. Features are extracted from each series and passed through the trained network to obtain a pairwise affinity matrix, whose entries represent the probabilities that each pair of time series belongs to the same cluster. In this framework, the neural network learns a mapping from feature representations to pairwise affinity scores, which encode the underlying clustering structure of the data. While one could alternatively design a network that directly outputs a clustering partition, learning an affinity structure provides greater flexibility, as it encodes richer information than a hard partition, including probabilistic relationships between series that can be used to construct soft partitions or quantify uncertainty. To obtain the final clustering partition, we apply standard graph-based methods, such as spectral clustering~\citep{Ng2001OnSC, Luxburg2007ATO} and community detection using the Louvain method~\citep{Blondel2008FastUO}, to this matrix. This step discretizes the learned affinities into cluster labels. Importantly, these procedures do not define the notion of similarity, but only convert the learned affinity structure into a clustering partition. In contrast to classical approaches, where the clustering algorithm determines the grouping through a predefined objective, here the affinity structure is learned in advance, and the role of the clustering algorithm is reduced to partition extraction. These graph-based methods operate on the same affinity matrix and provide flexibility for different application requirements, without altering the learned notion of similarity encoded by the network.

\begin{figure}
  	\centering
  	\includegraphics[width=0.58\textwidth]{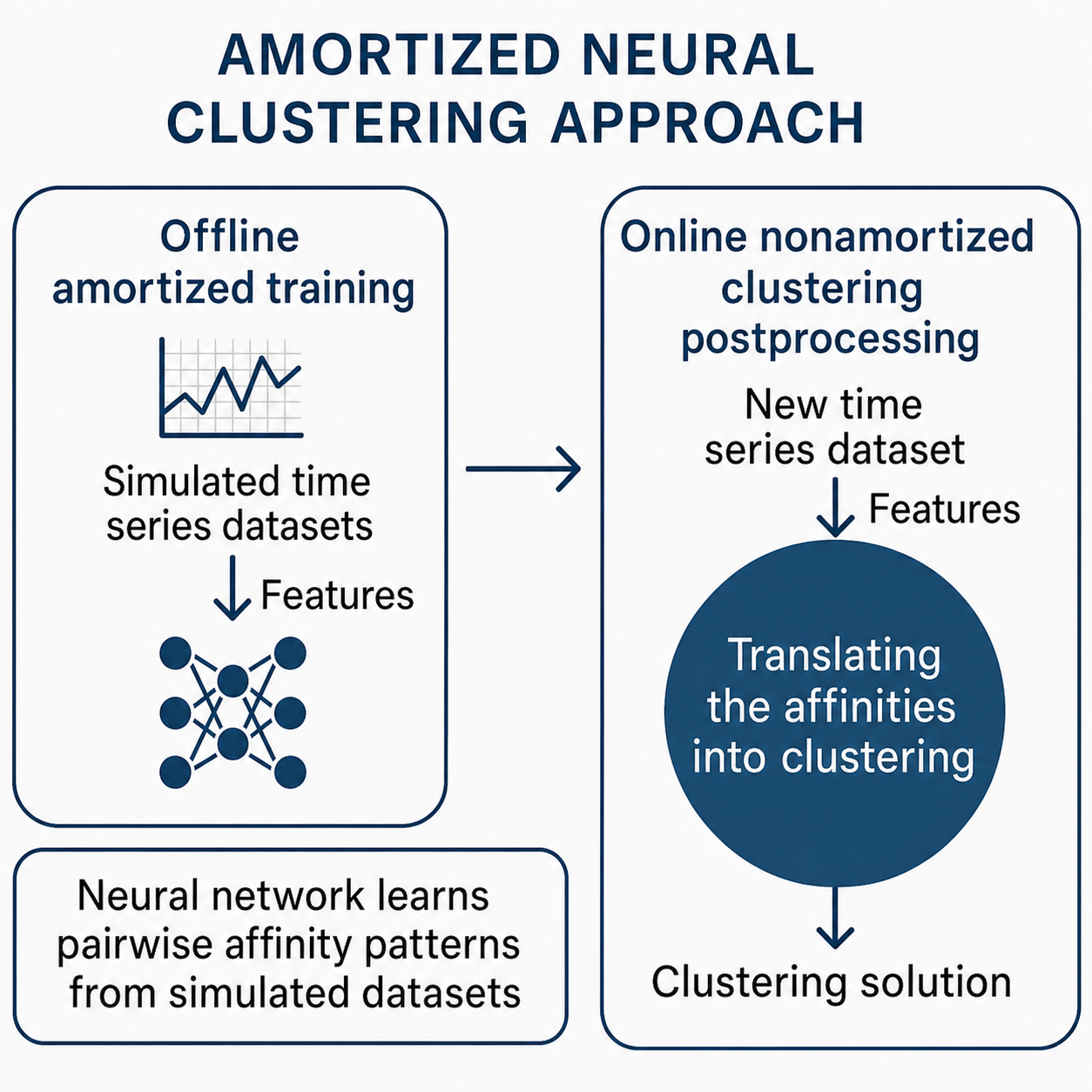}
  	\caption{Amortized neural clustering approach workflow. Offline amortized training uses simulated time series datasets for a neural network to learn pairwise affinity patterns. During online nonamortized clustering postprocessing, pairwise affinities for a new time series dataset are obtained with the trained network. These affinities are then translated into a final clustering partition.}
  	\label{flowchart}
   \end{figure}

    Figure \ref{flowchart} displays a summary of the proposed amortized neural clustering approach.

    \subsubsection{Evaluation and results}\label{subsubsectioner}
    
    To evaluate the proposed approach, the following procedure is adopted. Given values of \( n \), \( K \), and \( T \), a neural network is first trained on \( N=200000 \) datasets independently generated from Equation~\eqref{simm1}, as previously described. After training, 10000 new feature datasets are generated using the same simulation mechanism. Each dataset is then given as input to the trained network as indicated in Section \ref{subsubsectionna}, which produces a pairwise affinity matrix. From this matrix, clustering partitions are obtained using spectral clustering, applied using the known value of \( K \), and the Louvain community detection method, which does not require specifying the number of clusters. Notably, Louvain clustering is highly efficient due to its fast heuristic optimization and scalability to large, sparse graphs, aligning well with the amortized inference framework by deriving cluster structure directly from the learned affinities without additional tuning. 
    
\begin{figure}
  	\centering
  	\includegraphics[width=1\textwidth]{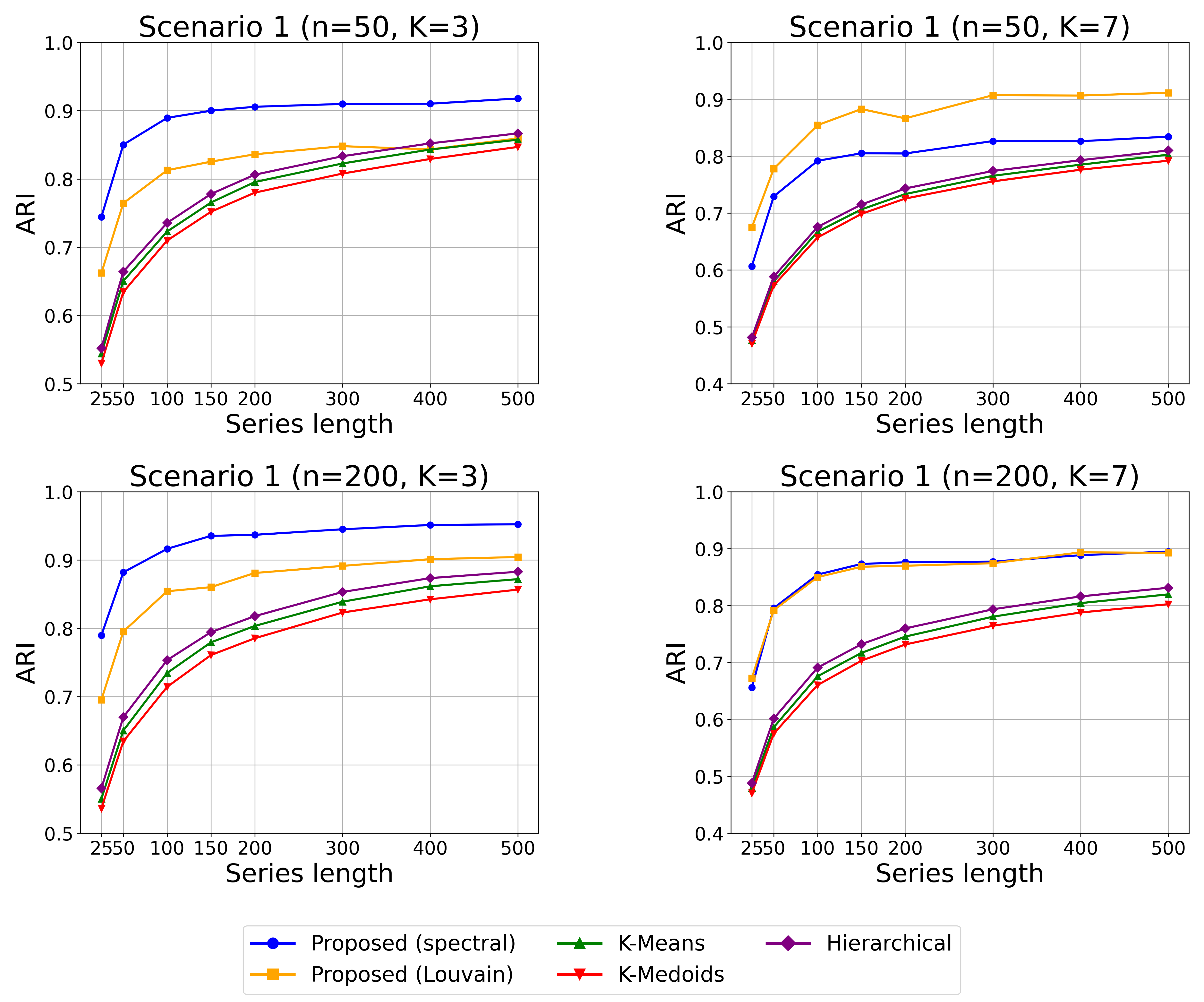}
  	\caption{Average clustering accuracy (ARI) for different methods as a function of the series length $T$ in Scenario 1. Each panel corresponds to specific values of $n$ and $K$.}
  	\label{scenario1}
  \end{figure}

    The resulting partitions are compared to the ground truth using the adjusted Rand index (ARI)~\citep{hubert1985comparing}, and the mean ARI across all 10000 datasets is reported as the performance metric. For comparison, standard time series clustering methods are also evaluated, in which clustering algorithms are applied directly to the feature datasets. Specifically, \(K\)-means (using 200 random initializations), \( K \)-medoids, and agglomerative hierarchical clustering with Ward’s linkage criterion are considered, which are widely used in the literature~\citep{Wang2007StructureBasedSF, lopez2021quantile}. Each algorithm is provided with the true value of \( K \) (for hierarchical clustering, the dendrogram is cut to yield \( K \) groups), and the average ARI is computed over the same collection of simulated datasets. Figure \ref{scenario1} shows the average ARI of the different methods as a function of the series length $T$, with each panel corresponding to specific values of $n$ and $K$.

   According to Figure~\ref{scenario1}, the proposed amortized neural clustering approach demonstrates notable improvements over standard methods for shorter and moderate values of~$T$. As the series length increases, the performance gap between the neural method and classical approaches narrows, with all methods achieving higher ARI values for longer series. However, this behavior is not uniform across all scenarios: in challenging settings featuring a large number of time series ($n=200$) and clusters ($K = 7$), the proposed neural framework retains a clear advantage, outperforming the traditional methods even for the largest values of~$T$ considered. This finding highlights the robustness and scalability of the neural method, particularly when the clustering task becomes more complex due to an increased number of groups. Among the two versions of the proposed technique, using spectral clustering to convert the pairwise affinity matrix into the clustering partition yields better results when three groups are considered ($K=3$), while using the Louvain method provides better or similar results compared to spectral clustering when seven groups are considered ($K=7$). Note that the results for $n=50$ and $K=7$ are particularly positive. In this setting, the proposed approach using the Louvain method achieves the best accuracy for all values of $T$, despite the fact that this method does not require prior knowledge of the true number of clusters, unlike the other procedures.

    \subsubsection{Variable number of time series and clusters}\label{subsubsectionvn}
    
    To examine the applicability and flexibility of the proposed simulation-based clustering approach, we introduce a more challenging generation scenario, which we will refer to as Scenario 2, designed to emulate complex and realistic data environments. Unlike the fixed-parameter setting in Scenario 1, where the number of time series (\(n\)) and clusters (\(K\)) are predetermined and constant, this new scenario treats \(n\) and \(K\) as random variables drawn from certain distributions. This probabilistic framework allows the network to learn an inference rule that generalizes across varying dataset sizes and cluster complexities, capturing a richer family of possible clustering universes. By exposing the neural model to a wider variety of configurations during training, it aims to robustly approximate an optimal decision rule capable of handling the uncertainties and heterogeneities inherent in many real-world time series datasets. This approach also mitigates the need for retraining or manual tuning when faced with datasets of differing sizes or numbers of clusters, thereby enhancing the adaptability and practical utility of the amortized clustering framework.
    
    The new setting considers the same type of generating processes and features as the previous one. In particular, Scenario 2 involves the following generating mechanism:
 	\begin{align}\label{simm2}
 		& n \sim U_{\text{D}}[10, 200], \notag \\
 		& \alpha_1 \sim \text{Exp}(1), \notag \\
 		& \pi_1, \ldots, \pi_n \mid \alpha_1 \sim \text{CRP}(\alpha_1), \notag \\
 		& K = \left| \left\{ \pi_i : i = 1, \ldots, n \right\} \right|, \\
 		& \kappa_j^k \sim U[-1,1], \quad j=1,2,3, \quad k=1, \ldots, K, \notag \\
 		& \sigma_k^2 \sim U[0.1, 2], \quad k=1, \ldots, K, \notag \\
 		& \text{Obtain AR coefficients } \boldsymbol{\phi}^k \text{ from } \boldsymbol{\kappa}^k \text{ via Durbin-Levinson recursion}, \quad k=1, \ldots, K, \notag \\
 		& \boldsymbol{X}_i \mid \pi_i, \boldsymbol{\phi}^{\pi_i}, \sigma_{\pi_i}^2 \sim \text{Gaussian AR}(3) \text{ with coefficients } \boldsymbol{\phi}^{\pi_i} \text{ and variance } \sigma_{\pi_i}^2, \quad i=1, \ldots, n, \notag
 	\end{align}
 
 \noindent
 where \(U_{\text{D}}[a, b]\) denotes the discrete uniform distribution over integers from \(a\) to \(b\), inclusive, and \(\mathrm{CRP}(\alpha_1)\) denotes the Chinese restaurant process (CRP) with concentration parameter \(\alpha_1\) \citep{pitman2006combinatorial}. The CRP is a distribution over partitions that allows for an unknown number of clusters. Intuitively, \(n\) customers (time series) sequentially choose to join existing clusters with probability proportional to their size or start a new cluster with probability proportional to \(\alpha_1\). This induces a random number of clusters \(K\), defined as the cardinality of the set of distinct cluster labels in \(\pi_1, \ldots, \pi_n\). The CRP has been employed in several works for generating clustering settings; see, e.g., \cite{pakman2020neuralclusteringprocesses}. In the above context, rejection sampling is used to discard CRP draws that produce a single cluster, ensuring that only partitions with \(K \ge 2\) are retained. The remaining steps mirror those of Scenario~1.
 
The neural network is trained in a manner similar to Scenario~1. However, in this case, each of the $N$ simulated datasets contains a varying number of time series ($n$) and clusters ($K$). The series length ($T$) is kept fixed during each training setting. This time, as the generating mechanism involves much richer clustering configurations, we set $N = 500000$. Evaluation is performed similarly to the previous scenario; that is, given a value of $T$, $10000$ time series collections are generated from Equation~\eqref{simm2}, and clustering partitions are obtained using both the proposed approach and the alternative methods. For evaluation purposes, the true value of $K$, which is now allowed to vary across datasets, is assumed to be known for both the version of the proposed method that uses spectral clustering and the competing algorithms. The average ARI is again used as the performance metric.

\begin{figure}
  	\centering
  	\includegraphics[width=0.75\textwidth]{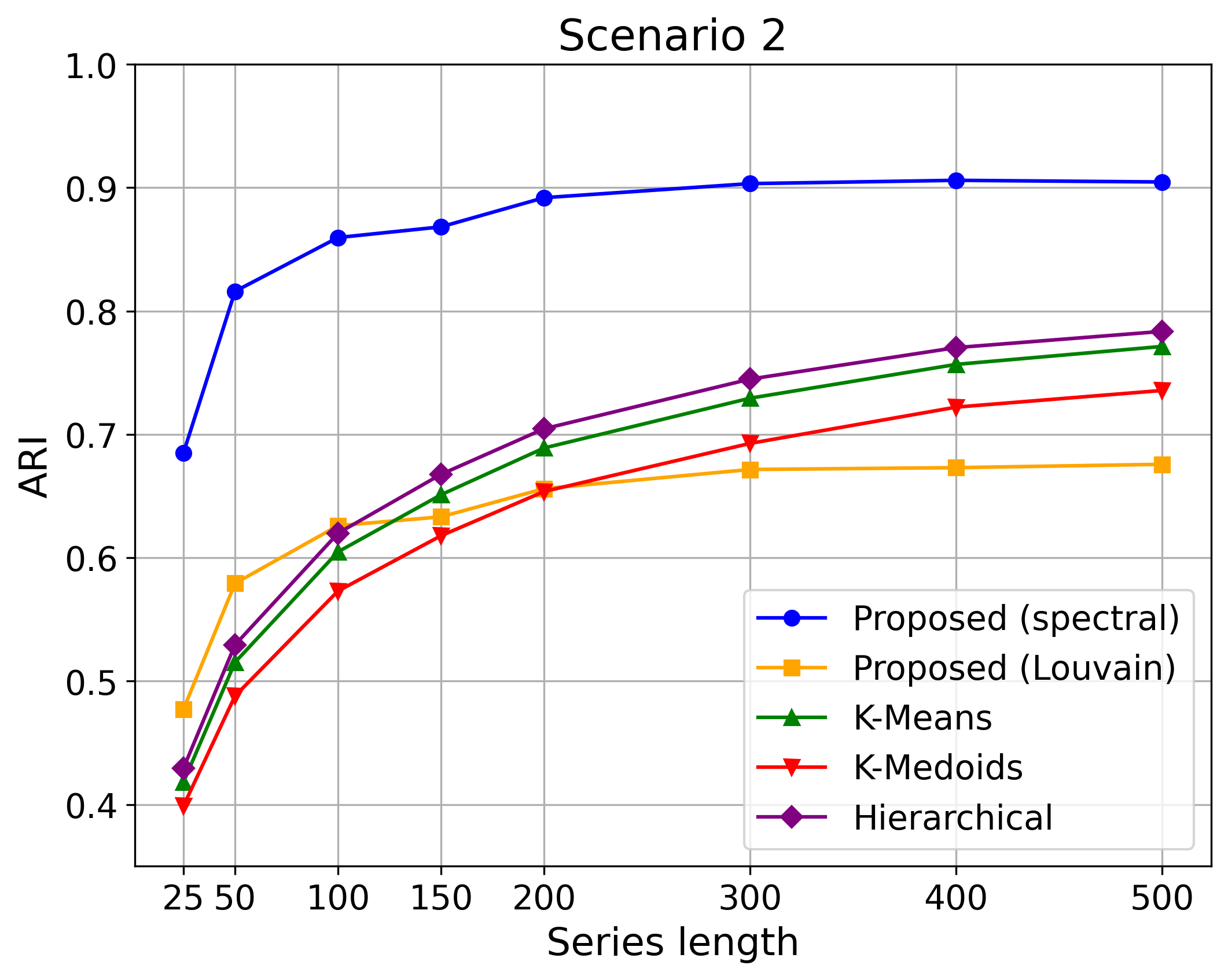}
  	\caption{Average clustering accuracy (ARI) for different methods as a function of the series length $T$ in Scenario 2.}
  	\label{scenario2}
  \end{figure}

Figure \ref{scenario2} presents the results for Scenario 2. The proposed simulation-based approach combined with spectral clustering achieves, by far, the highest clustering accuracy across all values of $T$. In contrast, the alternative version employing the Louvain method to construct the partition outperforms the classical clustering approaches for values of $T$ up to 100 but is outperformed by them beyond this point. These results are remarkable, as they demonstrate that within our simulation-based framework, the neural network effectively learns the underlying clustering structures in a complex generating universe such as the one defined in Equation~\eqref{simm2}. The results obtained by the Louvain-based version are noteworthy, as this approach, despite not being provided with the true number of clusters for each newly generated dataset, achieves performance that remains reasonably close to that of the traditional methods, and in some cases even surpasses them. In fact, to ensure a fair comparison between these methods and our Louvain-based technique, the traditional approaches would require an additional step to determine the true number of clusters for each dataset. Given the broad range of possible values of $K$ in the considered generative scheme, this extra step would likely reduce the performance of these methods substantially, while also increasing computational cost.

\begin{figure}
  	\centering
  	\includegraphics[width=1\textwidth]{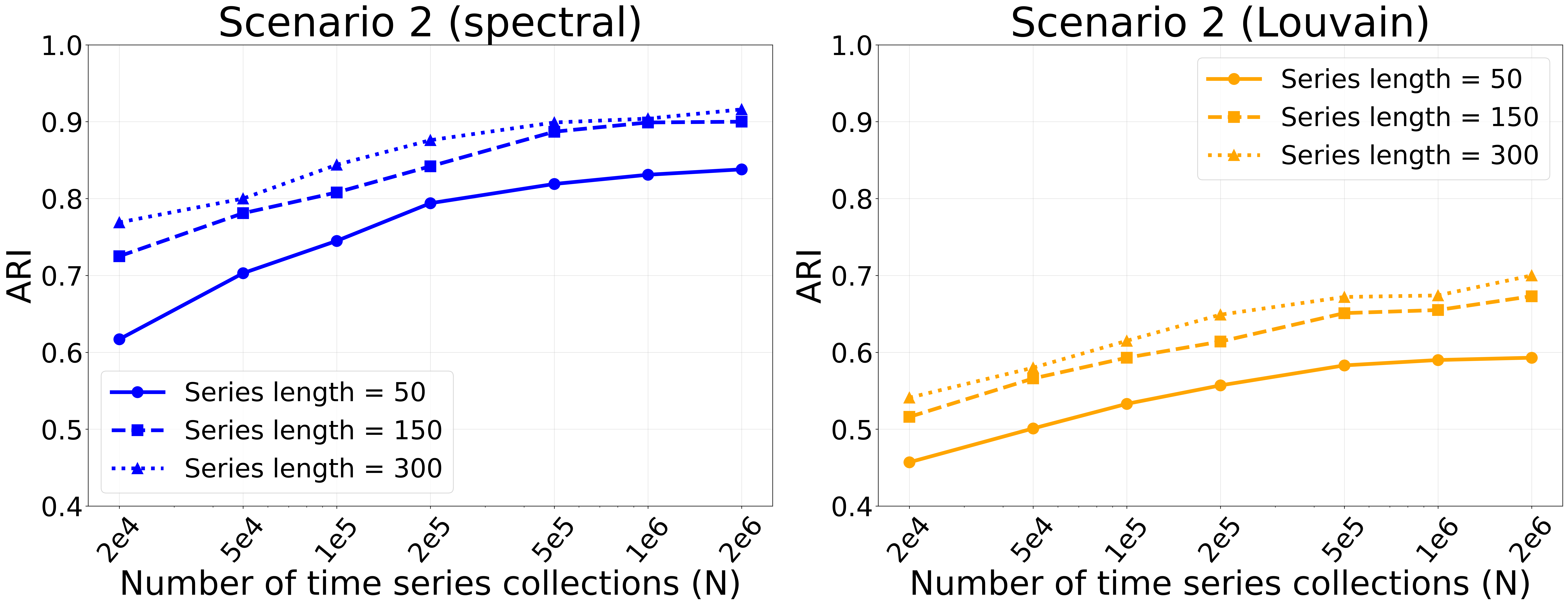}
  	\caption{Average clustering accuracy (ARI) for the proposed method using spectral clustering (left panel) and the Louvain technique (right panel) in Scenario 2, as a function of the number of time series datasets used for training the network ($N$), for three values of the series length ($T$).}
  	\label{scenario2extra}
\end{figure}

  To examine the underlying learning process of the neural network, we conduct an analysis analogous to the one described above, but focusing only on the proposed method. Here, the simulation mechanism in Equation~\eqref{simm2} is used again to generate collections of time series, focusing on lengths $T=50, 150, 300$. For each $T$, the neural network is trained using different values of $N$ ranging from 20000 to 2 million, and the proposed approach is evaluated on 10000 new collections as before. Figure~\ref{scenario2extra} shows the results, with clustering accuracy (ARI) plotted as a function of $N$. The left and right panels correspond to the proposed approach using spectral clustering and the Louvain method, respectively, with line types indicating different $T$ values. In all cases, increasing $N$ improves clustering accuracy on new datasets, with the most pronounced gains occurring between $N=20000$ and $N=500000$, followed by less marked improvements thereafter. This analysis underscores the importance of using a sufficient number of simulated collections to enable the network to approximate the optimal decision rule for the considered clustering context.

    \subsection{Clustering based on quantile autocorrelations}\label{subsectionsimulations2}

    Here, we evaluate the proposed amortized approach using quantile autocorrelations as features for time series clustering. The quantile autocorrelation function (QAF) generalizes the classical autocorrelation function by measuring dependence between quantile exceedances at different lags. In doing so, it captures not only linear relationships but also nonlinear dynamics and tail dependence, properties that traditional second-moment methods often miss. By leveraging QAF, we can obtain a richer summary of serial dependence structures, enabling more effective discrimination between time series generated by complex underlying processes. Several studies have demonstrated the benefits of using QAF-based and related features for time-series clustering~\citep{lafuente2016clustering, lafuente2020robust, alonso2020hierarchical, lopez2021quantile}, further motivating their use within our simulation-based framework.

  Specifically, for observations $X_{i,1}, \ldots, X_{i,T}$ from the time series $\bm{X}_i$, the population QAF at lag $l$ and quantile levels $\tau, \tau' \in (0,1)$ is defined as
\[
\rho_i^{(\tau, \tau')}(l)
= \mathrm{Corr}\left[ I(X_t^{(i)} \leq q_{i,\tau}),\; I(X_{t+l}^{(i)} \leq q_{i,\tau'}) \right],
\]
where $q_{i,\tau}$ denotes the $\tau$‑quantile of the strictly stationary process $\{X_t^{(i)},\, t \in \mathbb{Z}\}$, and $I(\cdot)$ is the indicator function.

A natural estimator of $\rho_i^{(\tau, \tau')}(l)$ is given by
\[
\widehat{\rho}_i^{(\tau, \tau')}(l)
= 
\frac{
\frac{1}{T} \sum_{t=1}^{T-l} 
I(X_{i,t} \leq \widehat{q}_{i,\tau})\,
I(X_{i,t+l} \leq \widehat{q}_{i,\tau'}) 
- \tau \tau'
}{
\sqrt{ \tau(1-\tau)\, \tau'(1-\tau') }
},
\]
where $\widehat{q}_{i,\tau}$ denotes the empirical $\tau$‑quantile of the series $\bm{X}_i$.

In what follows, these estimated QAF (computed over a grid of lags and quantile levels) constitute the feature vectors $\bm{\Psi}_i$ employed for clustering through the amortized neural framework.

We now focus on generalized autoregressive conditional heteroskedasticity (GARCH) processes as the underlying time series models. We choose these models because clustering GARCH-type time series is both an important problem in finance and a challenging task that allows for a stringent evaluation of our simulation-based framework \citep{d2016garch, tong2026cluster}. The GARCH($p$, $q$) family, introduced by \cite{bollerslev1986generalized} as an extension of the autoregressive conditional heteroskedasticity (ARCH) model of \cite{engle1982autoregressive}, captures stylized facts commonly observed in financial time series, such as time-varying volatility, volatility clustering, and heavy tails. For generality, in what follows we introduce the general GARCH($p$, $q$) model, although our experiments will focus on GARCH(1,1), the most common specification for financial time series \citep{tsay2005analysis, francq2019garch}.  

A zero-mean GARCH(\(p, q\)) process \(\{X_t, t \in \mathbb{Z}\}\) is defined by 
\begin{equation*}
    X_t = \sigma_t \varepsilon_t,
\end{equation*}
where $\{\varepsilon_t, t \in \mathbb{Z}\}$ is a sequence of i.i.d.\ random variables with zero mean and unit variance, and
\begin{equation*}
    \sigma_t^2 = \mathrm{Var}(X_t \mid \mathcal{F}_{t-1}) = \omega + \sum_{i=1}^p \alpha_i X_{t-i}^2 + \sum_{j=1}^q \beta_j \sigma_{t-j}^2,
\end{equation*}
where $\omega > 0$, $\alpha_i \geq 0$, $\beta_j \geq 0$ for $i=1,\ldots, p$, $j=1, \ldots, q$. Here, $\sigma_t^2$ represents the conditional variance of $X_t$ given the past, and $\mathcal{F}_{t-1}$ denotes the sigma-field generated by historical observations. The condition $\sum_{i=1}^p \alpha_i + \sum_{j=1}^q \beta_j < 1$ ensures second-order stationarity of the GARCH($p$, $q$) process. For strict stationarity, the condition is more complex in general \citep{bougerol1992stationarity}, but it simplifies for the GARCH(1,1) model to $\mathbb{E}\big[\log(\alpha \varepsilon_t^2 + \beta)\big] < 0$, where the subscript has been dropped for notational convenience.

Throughout this paper, we consider GARCH(1,1) models where $\varepsilon_t$ follows a standardized Student-$t$ distribution with $\nu>2$ degrees of freedom, denoted as $t_{\nu}$. Thus, the parameters are $\omega$, $\alpha$, $\beta$, and $\nu$. This choice accommodates the heavy tails and kurtosis typically observed in financial time series, providing a more realistic representation than Gaussian innovations \citep{ardia2010bayesian}.    

We note that, although the use of QAF-based features may not appear, at first glance, to be the most suitable choice for discriminating between different GARCH processes (since alternative features, such as estimated coefficients obtained via maximum likelihood, might seem more appropriate), an empirical justification for employing QAF-based features in the context of the GARCH(1,1) models considered in this section is provided in the Supplement. In particular, we show that, in a clustering setting similar to the one considered here, certain QAF-based features can be at least as accurate as the classical estimated GARCH(1,1) coefficients in detecting the underlying clustering partition. Moreover, using QAF-based features versus model coefficients is advantageous for clustering, since the latter are clearly sensitive to model misspecification and can lead to poor clustering solutions if some series depart from the assumed true model, while the former can still cluster the time series meaningfully in terms of dependence structures under strict stationarity. In any case, the proposed framework can be readily extended to features based on estimated GARCH model parameters.

For the experiments in this section, we consider a variable number of clusters and time series. To create cluster-specific generative models, for cluster $k$, $k = 1, \ldots, K$, we first randomly sample the corresponding GARCH(1,1) coefficients $\bm{\theta}_k = (\omega_k, \alpha_k, \beta_k)$. For each time series, $\nu$ is then sampled from a specific distribution and may vary within clusters. Specifically, the generating mechanism, referred to as Scenario 3, is as follows:
 	\begin{align}\label{simm3}
 		& n \sim U_{\text{D}}[10, 100], \notag \\
 		& \alpha_1 \sim \text{Exp}(0.5), \notag \\
 		& \pi_1, \ldots, \pi_n \mid \alpha_1 \sim \text{CRP}(\alpha_1), \notag \\
 		& K = \left| \left\{ \pi_i : i = 1, \ldots, n \right\} \right|, \notag \\
 		& \omega_k \sim U[10^{-6}, 10^{-4}], \quad k=1, \ldots, K, \\
 		& \alpha_k \sim U[0.01, 0.30], \quad k=1, \ldots, K,  \notag \\
        & \beta_k \sim U[0.70, 1-\alpha_k], \quad k=1, \ldots, K, \notag \\
 		& \text{For the $i$th time series, generate } \nu_i \sim U(\{3, 4, 5, 6, 7, 8, 10000\}), \notag\\
 		& \boldsymbol{X}_i \mid \pi_i, \boldsymbol{\theta}_{\pi_i}, \nu_i \sim \text{GARCH}(1,1) \text{ with coefficients } \boldsymbol{\theta}_{\pi_i} \text{ and } \varepsilon_t \sim t_{\nu_i}, \quad i=1, \ldots, n, \notag
 	\end{align}
\noindent where $U(\cdot)$ denotes the discrete uniform distribution on the elements of the input set. the Note that the GARCH(1,1) coefficients $\omega_k$, $\alpha_k$, and $\beta_k$ are sampled from uniform distributions with ranges reflecting values commonly estimated from real-world financial time series. Numerous empirical studies report typical $\alpha \in [0.05, 0.20]$ and $\beta \in [0.80, 0.95]$ for financial returns across different contexts~\citep{tsay2005analysis, francq2019garch}, with high $\beta$ capturing volatility persistence. We slightly extend these to encompass broader clustering problems, including slightly extreme volatility regimes. For the degrees of freedom, $\nu_i$ spans heavy‑tailed ($\nu_i = 3$) to  fairly Gaussian ($\nu_i = 8$) and virtually Gaussian ($\nu_i = 10000$) innovations, realistic for financial data where innovations can be Gaussian but frequently exhibit moderate heavy tails~\citep{ardia2010bayesian}. 

Draws are constructed so that strict stationarity holds in all cases. Note that the parameters are generated to satisfy $\alpha_k + \beta_k < 1$, a condition frequently assumed in the applied literature on GARCH(1,1) models. Under the standard assumption $\mathbb{E}[\varepsilon_t^2]=1$, Jensen's inequality implies $\mathbb{E}[\log(\alpha_k \varepsilon_t^2 + \beta_k)] \le \log(\mathbb{E}[\alpha_k \varepsilon_t^2 + \beta_k]) = \log(\alpha_k + \beta_k) < 0$. Therefore, the strict stationarity condition is automatically satisfied for all simulated processes, ensuring that the population quantities underlying the QAF-based features exist.

Critically, the above scheme creates a meaningful clustering task: group series by volatility dynamics (GARCH parameters driving persistence and leverage) despite heterogeneous noise distributions within clusters. This mirrors real financial applications; e.g., clustering assets by volatility regimes while allowing varying tail risks, and tests whether QAF features robustly recover volatility structure independent of marginal tails.

The neural network is trained similarly to Scenario 2, with the series length fixed at $T=5000$ (as estimating GARCH‑type models typically requires long samples) and $N=200000$ time series collections for training. In this setting, feature extraction is performed by computing the estimated QAF for certain sets of probability levels and lags. In particular, for the $i$th time series, we consider the collection of features $\widehat{\bm{\rho}}^{(i)}=\bigl\{\widehat{\rho}_i^{(\tau,\tau')}(l):\,\tau,\tau' \in \mathcal{T},\; l \in \mathcal{L}\bigr\}$, where $\mathcal{T}=\{0.1, 0.5, 0.9\}$ and $\mathcal{L}=\{1, 2, 3\}$. A justification for employing such sets of probability levels and lags is provided in the Supplement. The resulting feature vector for the $i$th series, $\bm{\Psi}_i$, is then constructed by concatenating all elements in $\widehat{\bm{\rho}}^{(i)}$. Evaluation follows the protocol of the previous scenarios.

\begin{figure}
  	\centering
  	\includegraphics[width=0.75\textwidth]{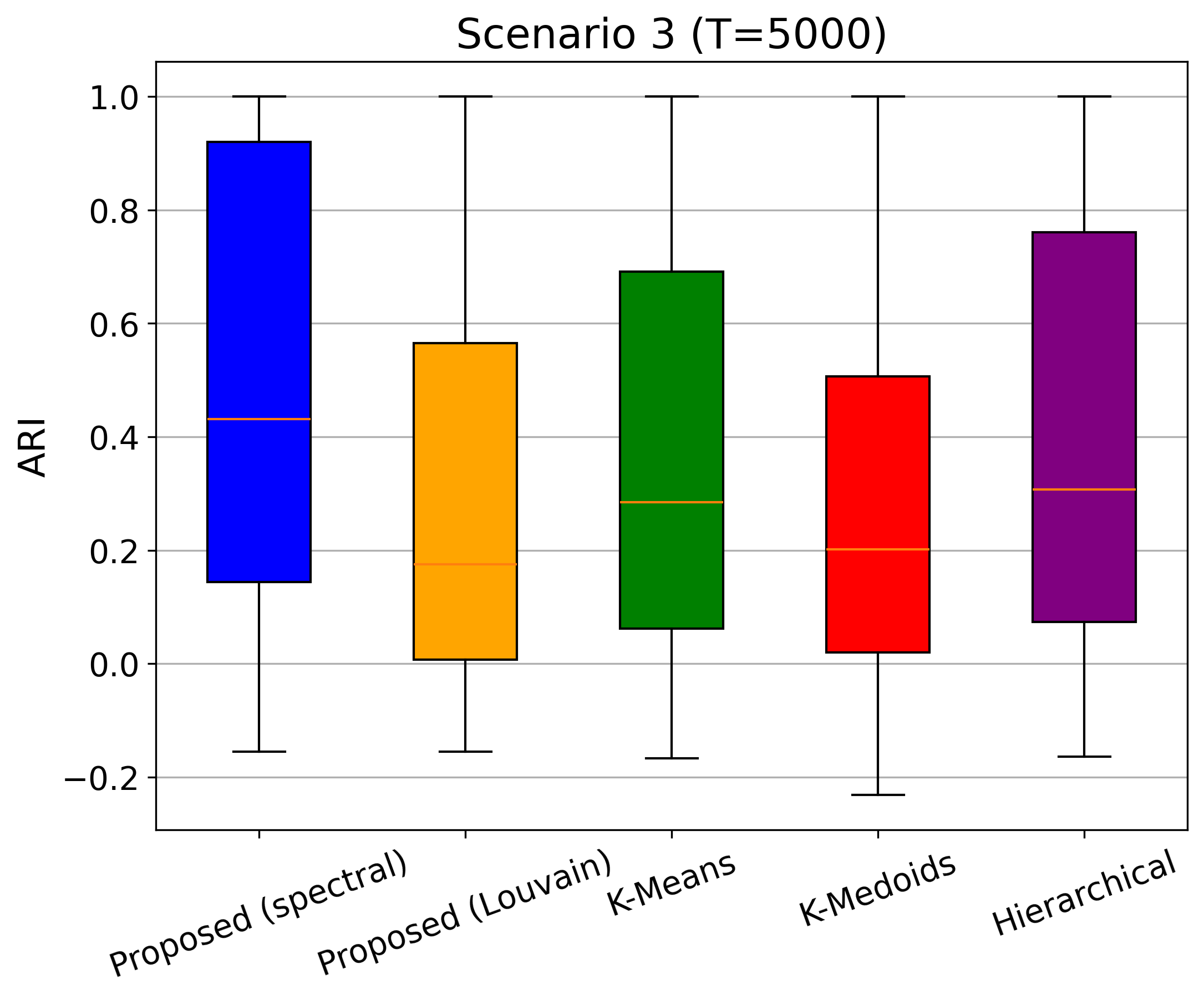}
  	\caption{Distribution of clustering accuracy (ARI) across the different methods in Scenario~3 for series of length \(T=5000\).}
  	\label{scenario3}
  \end{figure}

Figure~\ref{scenario3} displays boxplots of clustering accuracy for the various methods in Scenario 3. The proposed approach using spectral clustering achieves the highest accuracy, suggesting that the network effectively learns to differentiate among GARCH(1,1) processes using QAF-based features. The Louvain-based version performs rather worse, but it remains not far from the alternative methods, despite not requiring the true number of clusters as an input. In this setting, clustering accuracy remains modest for all methods, even with time series of length $T=5000$. This limitation may stem from two factors: (i) the complexity of GARCH(1,1) processes, which are challenging to estimate accurately even with long series (specially under heavy tails), and (ii) the ranges selected for the simulation parameters $\alpha_k$ and $\beta_k$, which can generate datasets with similar volatility dynamics across different clusters, making them difficult to distinguish even when they are defined as distinct for evaluation. Table~\ref{tables3} reports the corresponding ARI means and medians for each method.

\begin{table}[h]
\centering
\begin{tabular}{lcc}
\toprule
Method & Mean ARI & Median ARI \\
\midrule
Proposed (spectral)   & 0.490 & 0.431 \\
Proposed (Louvain)    & 0.311 & 0.175 \\
$K$-Means               & 0.388  & 0.285 \\
$K$-Medoids             & 0.312 & 0.202  \\
Hierarchical          & 0.409  & 0.307  \\
\bottomrule
\end{tabular}
\caption{Mean and median clustering accuracy (ARI) for the different methods in Scenario~3 for series of length \(T=5000\).}
\label{tables3}
\end{table}

To further analyze the performance of the proposed framework, we conduct detailed pairwise comparisons between its two versions and each alternative procedure. For each pair, we report the proportion of replicates where one outperforms the other, as well as the proportion of ties. We also provide the $p$-value from the corresponding Wilcoxon signed-rank test. Results in Table~\ref{tablecomp} confirm the strong performance of the spectral-based version of the proposed framework.

\begin{table}[h]
\centering
\begin{tabular}{l c c c c}
\toprule
Pair & Proposed $>$ Alternative & Alternative $>$ Proposed & Ties & $p$-value \\
\midrule
Spectral vs. $K$-means & 0.525 & 0.307 & 0.168 & 0.000 \\
Spectral vs. $K$-medoids & 0.631 & 0.253 & 0.116 & 0.000 \\
Spectral vs. Hierarchical & 0.489 & 0.317 & 0.194 & 0.000 \\
Louvain vs. $K$-means & 0.443 & 0.508 & 0.049 & 0.000 \\
Louvain vs. $K$-medoids & 0.556 & 0.399 & 0.045 & 0.000 \\
Louvain vs. Hierarchical & 0.439 & 0.506 & 0.055 & 0.000 \\
\bottomrule
\end{tabular}
\caption{Comparison of clustering methods in Scenario 3 for series of length \(T=5000\).}
\label{tablecomp}
\end{table}
 
\subsection{Additional analyses}\label{subsectionsimulations3}

In the previous sections, the method was evaluated under the same generative mechanisms used to train the network. However, in practice, the true data generating processes are rarely known with certainty, and it is therefore important to examine how the method behaves when the clustering models are misspecified or differ from those assumed in training. This type of robustness analysis is particularly relevant in the present setting, as it mirrors the broader concerns of simulation-based inference methods, which may produce misleading results under model misspecification \citep{huang2023learning} and thus motivates a careful assessment of performance beyond the well-specified case.

To assess the robustness of the method, we train the neural network on AR(3) time series generated as in Scenario~\ref{simm2}, that is, according to Equation~\eqref{simm2}, but using QAF-based features computed with \(\mathcal{T}=\{0.1,0.5,0.9\}\) and \(\mathcal{L}=\{1,2,3\}\), as in Scenario~\ref{simm3}. We consider time series of length \(T=300\) and generate a total of \(N=500000\) time series collections for training. After training, we evaluate clustering accuracy on 10000 new datasets, where the underlying clusters are, with some probability, given by processes outside the classical autoregressive family. In particular, we consider self-exciting threshold autoregressive (SETAR) models \citep{de1998forecasting}. In particular, SETAR models with two regimes where the process
in each regime is AR(1), frequently denoted as SETAR(2,1,1) models, introduced below.

A SETAR(2,1,1) process $\{X_t, t \in \mathbb{Z}\}$ is defined by the recursion  
\[
X_t = 
\begin{cases} 
\phi_{1,1} X_{t-1} + \varepsilon_t^{(1)} & \text{if } X_{t-1} \le r, \\
\phi_{2,1} X_{t-1} + \varepsilon_t^{(2)} & \text{if } X_{t-1} > r,
\end{cases}
\]  
where $\phi_{j,1}$ for $j=1,2$ are the AR(1) coefficients in regime $j$, $r \in \mathbb{R}$ is the threshold parameter, and $\{\varepsilon_t^{(j)}, t \in \mathbb{Z}\}$ is white noise with zero mean and variance $\sigma_j^2$. Hereafter, we assume for simplicity that the noise is standard Gaussian. A sufficient condition for the strict stationarity of the process is $\max_{j=1,2} |\phi_{j,1}| < 1$ \citep{tong2012threshold}.

The generating mechanism for the evaluation, referred to as Scenario~4, is defined as follows: with probability \(0.5\), we generate a time series collection according to Equation~\eqref{simm2}; otherwise, with probability \(0.5\), we generate a collection from a SETAR-based framework. In the latter case, we first sample the parameters \(\bm{\theta}_k = (\phi_{1,1}^{(k)}, \phi_{2,1}^{(k)}, r_k)\) for each cluster \(k\), and then generate each time series from the corresponding SETAR(2,1,1) model. The scheme used for the SETAR collections is as follows:
\begin{align}\label{simm4}
 		& n \sim U_{\text{D}}[10, 200], \notag \\
 		& \alpha_1 \sim \text{Exp}(1), \notag \\
 		& \pi_1, \ldots, \pi_n \mid \alpha_1 \sim \text{CRP}(\alpha_1), \notag \\
 		& K = \left| \left\{ \pi_i : i = 1, \ldots, n \right\} \right|, \\
 		& \phi_{j,1}^{(k)} \sim U[-1,1], \quad j=1,2, \quad k=1, \ldots, K, \notag \\
        & r_k \sim U[-0.75,0.75], \quad k=1, \ldots, K, \notag \\
 		& \boldsymbol{X}_i \mid \pi_i, \boldsymbol{\theta}_{\pi_i} \sim \text{SETAR}(2,1,1) \text{ with coefficients } \boldsymbol{\theta}_{\pi_i}, \quad i=1, \ldots, n. \notag
 \end{align}

By evaluating the neural network on time series collections generated as described above, we assess the robustness of the proposed framework to model misspecification, since the data generating mechanism is misspecified in half of the cases. In those instances, the generating processes shift from linear AR(3) dynamics to nonlinear SETAR dynamics, and the dependence patterns change from a third-order autoregressive form to a regime-dependent first-order structure. Although SETAR models differ from the training models, they still share some linear structure across regimes, which makes the misspecification moderate rather than extreme. When the test series depart substantially from the training assumptions, however, performance may deteriorate sharply. For this reason, ensuring sufficient similarity between training and application settings remains the analyst's responsibility, typically guided by domain knowledge and exploratory analysis.

\begin{figure}
  	\centering
  	\includegraphics[width=0.75\textwidth]{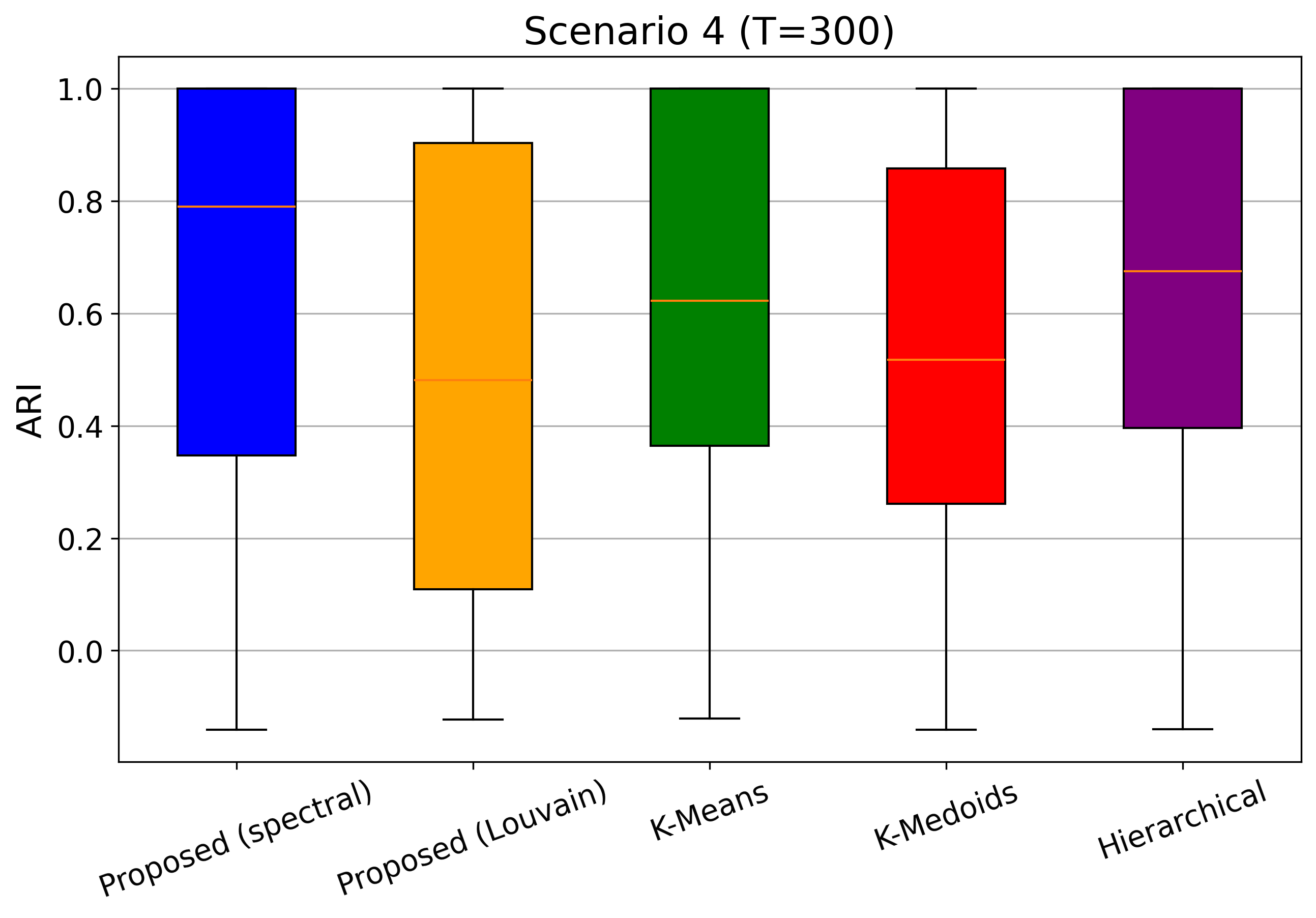}
  	\caption{Distribution of clustering accuracy (ARI) across the different methods in Scenario~4 for series of length \(T=300\).}
  	\label{scenario4}
\end{figure}

Results for Scenario~4 are presented in Figure~\ref{scenario4} as boxplots, while Table~\ref{tables4} reports the corresponding means and medians for each method. We observe that the proposed approach with spectral clustering is the best performing method, which indicates that the framework remains robust under moderate deviations from the training models. The Louvain-based version also achieves reasonable accuracy, especially given that it must infer the number of clusters automatically. In addition, the strong performance of the five methods in this setting further supports the suitability of QAF as a powerful feature for clustering time series based on dependence structures \citep{lafuente2016clustering, lafuente2020robust}.

\begin{table}[h]
\centering
\begin{tabular}{lcc}
\toprule
Method & Mean ARI & Median ARI \\
\midrule
Proposed (spectral)   & 0.665 & 0.790 \\
Proposed (Louvain)    & 0.503 & 0.481 \\
$K$-Means               & 0.621 & 0.623 \\
$K$-Medoids             & 0.531 & 0.517 \\
Hierarchical          & 0.646 & 0.675 \\
\bottomrule
\end{tabular}
\caption{Mean and median clustering accuracy (ARI) for the different methods in Scenario~4 for series of length \(T=300\).}
\label{tables4}
\end{table}

\section{Application: Clustering time series of stock returns}
\label{sectionapplication}

\begin{figure}
  	\centering
  	\includegraphics[width=1\textwidth]{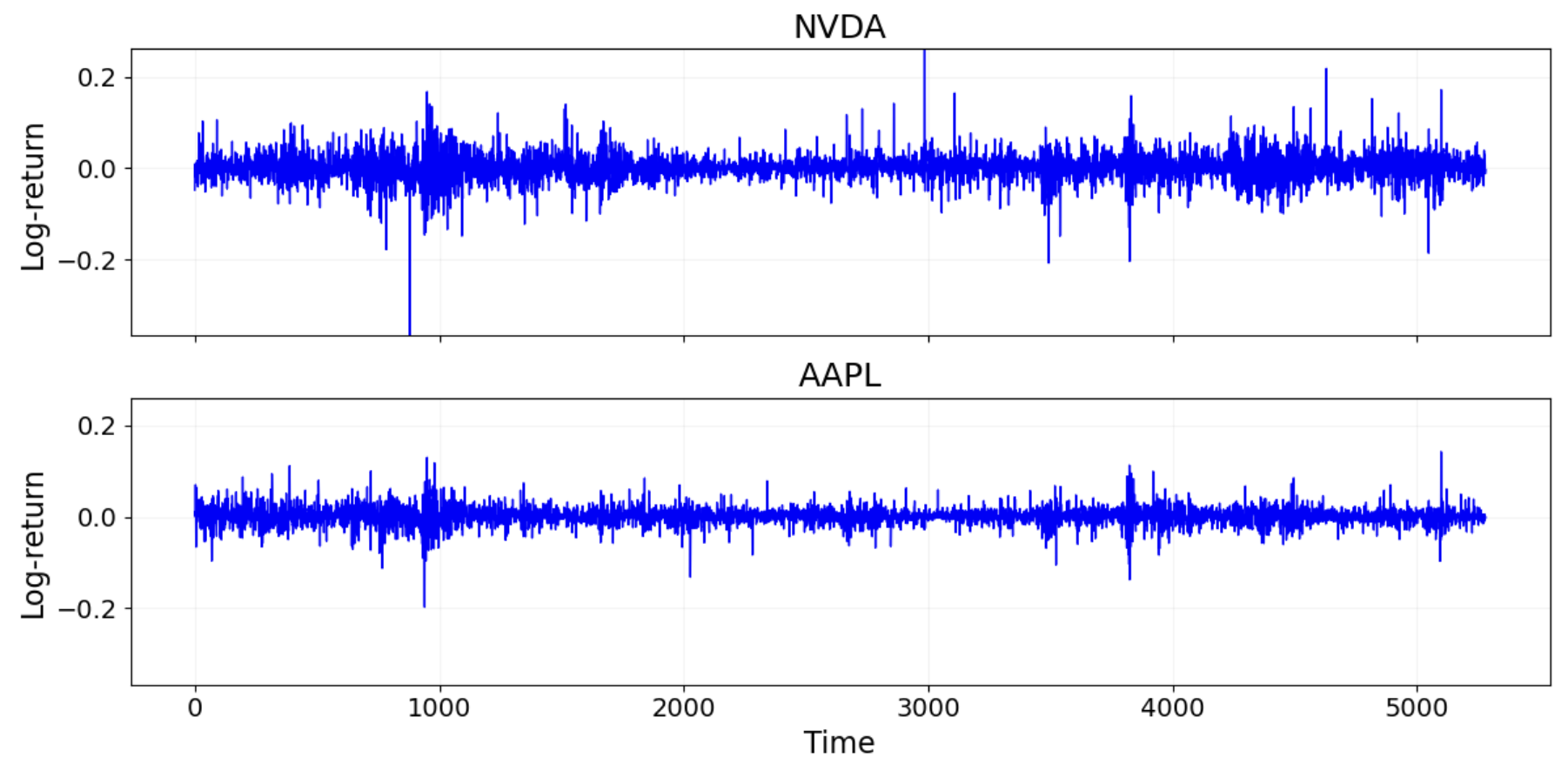}
  	\caption{Time series of log-returns of companies NVDA (top panel) and APPL (bottom panel).}
  	\label{returns}
\end{figure}

We focus on clustering financial time series for several S\&P 500 companies. In particular, we consider the daily adjusted closing prices of 50 of the most important companies currently included in the S\&P 500 index, over the period from January 1, 2005 to December 31, 2025. The time series are extracted using the \proglang{R} package \textbf{quantmod} \citep{quantmod}, which provides data for 5282 trading days within the considered period. As is standard in financial time series analysis, we transform prices into log-returns. If \(P_t\) denotes the adjusted closing price of one company at time \(t\), then the corresponding log-return is
\(r_t = \log\!\left(\frac{P_t}{P_{t-1}}\right)\).
This transformation is useful for several reasons \citep{tsay2005analysis}. First, it expresses changes on a relative scale, which makes assets with different price levels directly comparable. Second, log-returns are additive over time, so multi-period changes can be accumulated by summing one-period log-returns. Third, prices are usually nonstationary, whereas returns tend to be much closer to stationarity, making them better suited for statistical modeling. Therefore, the data collection under analysis consists of 50 time series, each of length \(T = 5281\) after transformation. Table~\ref{tablecompanies} summarizes the companies considered in the time series collection by listing their corresponding S\&P 500 ticker symbols. Log-returns for companies NVDA and AAPL are shown in Figure \ref{returns}. 

\begin{table}[h]
\centering
\begin{tabular}{llllllllll}
\toprule
NVDA & AAPL & MSFT & AMZN & GOOGL & GOOG & WMT & BRK-B & JPM & LLY \\
XOM & AMD & MU & JNJ & ORCL & COST & INTC & NFLX & CAT & BAC \\
CVX & CSCO & PG & HD & LRCX & AMAT & KO & UNH & MS & GE \\
MRK & GS & KLAC & TXN & WFC & LIN & RTX & C & IBM & AXP \\
MCD & PEP & NEE & ADI & VZ & AMGN & APH & T & BA & DIS \\
\bottomrule
\end{tabular}
\caption{Companies considered in the collection of time series of log-returns.}
\label{tablecompanies}
\end{table}

In our context, the goal is to cluster financial time series according to their volatility dynamics. Since daily return series often exhibit persistent conditional heteroskedasticity, the GARCH(1,1) model is a natural choice in this setting \citep{tsay2005analysis, francq2019garch}. Its popularity comes from its parsimony and flexibility: with only three parameters, it captures persistence and mean reversion while remaining easy to interpret, which is why it is often treated as a workhorse model for financial returns. We therefore assume a GARCH(1,1) structure for each series and cluster the data based on the resulting volatility features.

To perform clustering through our simulation-based framework, we train a neural network using a simulation scheme analogous to Equation~\eqref{simm3} for Scenario 3, but fixing the number of time series to \(n=50\) and the series length to \(T=5281\). In this way, the simulation setting matches the structure of the time series collection under analysis. The network is trained using QAF-based features computed on the quantile levels \(\mathcal{T}=\{0.1,0.5,0.9\}\) and the lags \(\mathcal{L}=\{1,2,3\}\). These choices are motivated in Sections~\ref{subsectionsimulations2}, as well as in the Supplement. We generate \(N=200000\) time series collections for training.

\begin{figure}
  	\centering
  	\includegraphics[width=1\textwidth]{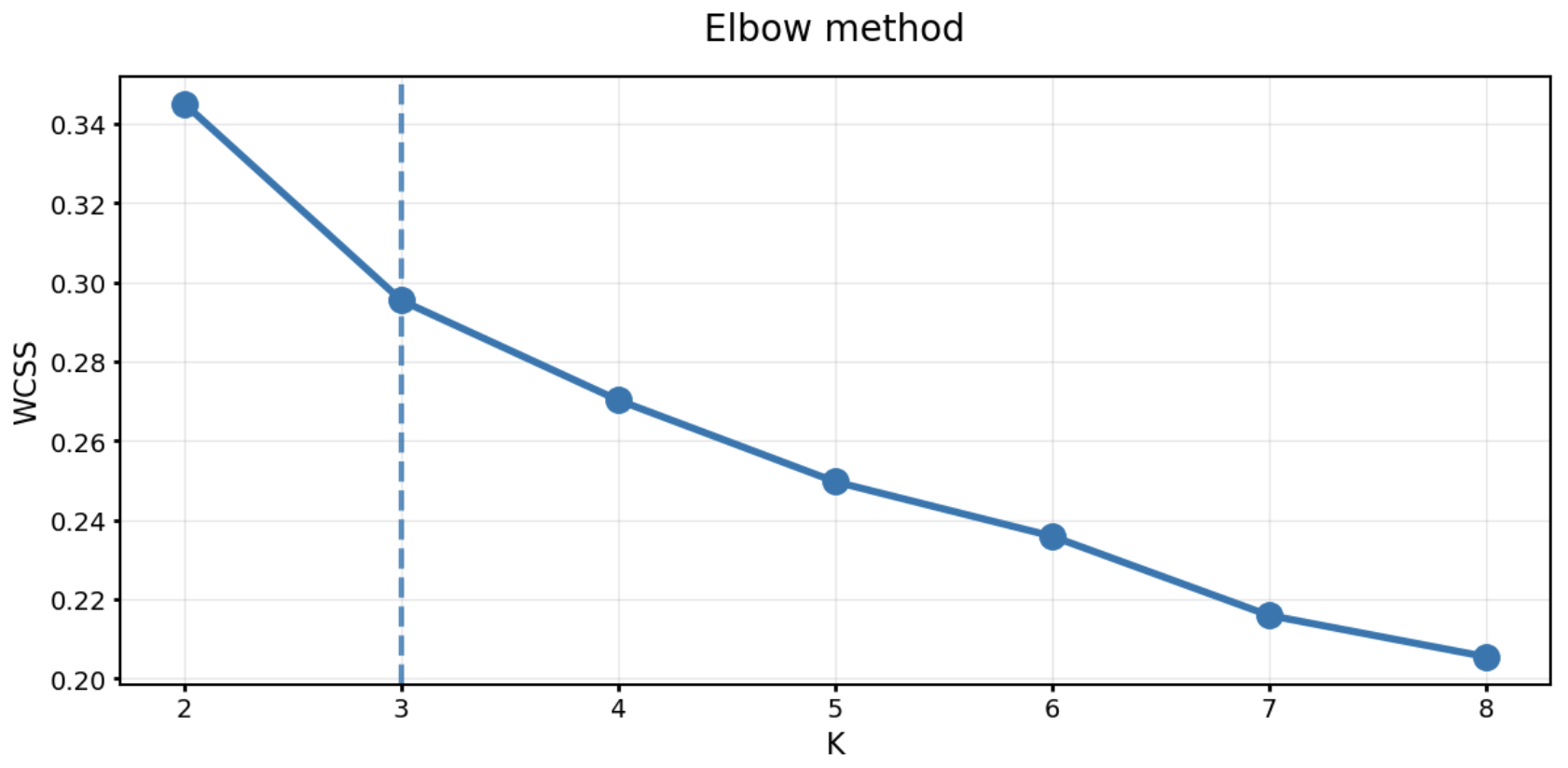}
  	\caption{Elbow plot for the time series of log-returns based on the clustering solution given by the spectral version of the proposed approach and the QAF features vectors.}
  	\label{elbow}
\end{figure}

Once the neural network is trained, we extract the QAF-based features from the 50 financial time series and compute the pairwise affinity matrix via a single forward pass. Given that the spectral-based approach is expected to clearly outperform the Louvain-based version in this setting (see Section \ref{subsectionsimulations2}), we use the former to obtain the clustering partition from the affinity matrix. Since spectral clustering requires a prespecified number of clusters $K$, we determine this parameter using the elbow method \citep{thorndike1953belongs}. Specifically, we compute the clustering solution and the associated within-cluster sum of squares (WCSS) based on the QAF feature vectors for each $K \in \{1, \dots, 8\}$. Figure \ref{elbow} displays the WCSS as a function of $K$; the largest decrease occurs between $K=2$ and $K=3$, making $K=3$ a reasonable choice. The resulting partition into three groups is described in Table \ref{tablecs}. Notably, Cluster 1 comprises 39 series, while the remaining two groups are significantly smaller, identifying one prevalent volatility pattern alongside two distinct dynamics. This structure is consistent with our training regime: by training on thousands of simulated 50-series datasets with frequently unequal cluster sizes due to the CRP, the network learns to robustly identify such heterogeneous partitions.

\begin{table}[h]
\centering
\begin{tabularx}{\textwidth}{l l X}
\hline
Cluster & Size & Companies \\
\hline
1 & 39 & NVDA, AAPL, MSFT, AMZN, WMT, BRK-B, LLY, XOM, MU, JNJ, ORCL, COST, INTC, CAT, CVX, CSCO, PG, HD, LRCX, AMAT, KO, UNH, MS, GE, MRK, GS, KLAC, TXN, LIN, RTX, IBM, MCD, PEP, NEE, VZ, APH, T, BA, DIS \\
2 & 5 & JPM, BAC, WFC, C, AXP \\
3 & 6 & GOOGL, GOOG, AMD, NFLX, ADI, AMGN \\
\hline
\end{tabularx}
\caption{Clustering solution obtained using the proposed approach in combination with the spectral method.}
\label{tablecs}
\end{table}

\begin{figure}
  	\centering
  	\includegraphics[width=1\textwidth]{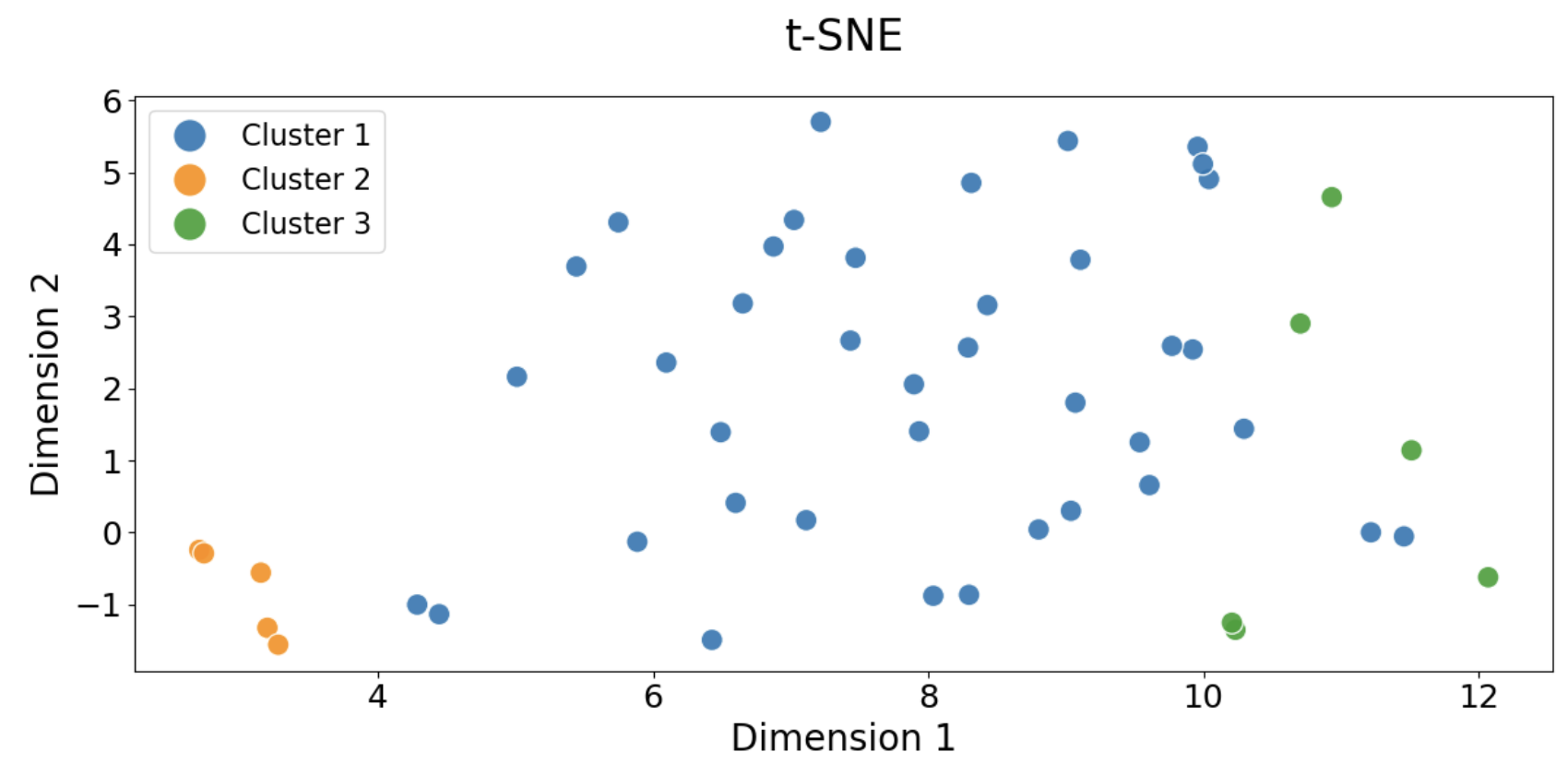}
  	\caption{Two-dimensional t-SNE plot for the time series of log-returns based on the corresponding QAF features vectors.}
  	\label{tsne}
\end{figure}

To validate the clustering solution, we apply t-distributed stochastic neighbor embedding (t-SNE) dimensionality reduction \citep{van2008visualizing} to the QAF feature vectors. This technique projects these feature vectors into a two-dimensional space, effectively preserving local neighborhood structures and highlighting cluster separation. The resulting t-SNE plot is shown in Figure~\ref{tsne}, where each color corresponds to a group in the partition of Table \ref{tablecs}. The plot reveals a rather clear separation between the three clusters, particularly along the first dimension, confirming that the groups possess distinct dependence structures as measured by QAF, which in turn reflect differing volatility patterns (see Section~\ref{subsectionsimulations2}).

This application illustrates how the proposed framework can be readily adapted to the time series collection under study once a specific clustering objective has been defined. In this case, we used the method to identify volatility patterns in financial time series, but the framework can naturally support other tasks as well. For example, if the goal is forecasting and the series are known to arise from a particular model class, the framework can be run with a suitable simulation universe and appropriate features, such as model-based features or more flexible alternatives like those derived from QAF. Then, a model can be fitted within each cluster to perform groupwise forecasting, with the expectation of improving upon the forecasts obtained by fitting a single model to each time series \citep{montero2021principles, lopez2025time}.
   
\section{Discussion}\label{sectiondiscussion}

This paper introduces a novel amortized neural inference framework for feature-based time series clustering, addressing longstanding challenges in traditional approaches. By training neural networks to approximate optimal partitioning rules from simulated data, the method reduces the need for algorithm selection (e.g., $K$-means, $K$-medoids, hierarchical clustering), explicit objective functions, and heuristic optimization. Empirical results for AR(3) and GARCH(1,1) processes show consistent superiority or competitiveness against classical alternatives, particularly for complex cluster structures. The pairwise affinity learning paradigm, based on permutation-invariant architectures, enables scalable inference via a single forward pass, making it well-suited for large-scale or streaming applications.

A key strength of our approach is its data-driven automation and adaptability. Unlike conventional methods, which may suffer from local minima or mismatched cluster assumptions, the amortized approach learns an affinity structure directly from diverse simulations. This yields meaningful partitions even when combined with simple graph-based procedures such as spectral clustering or Louvain detection, without requiring the true number of clusters in the latter case. Flexibility in feature design (from classical autocorrelations to quantile-based measures capturing nonlinear and tail dependence) allows adaptation to domain-specific dynamics, as illustrated in the GARCH(1,1) setting. Training costs are incurred upfront, enabling fast clustering of new datasets, which is particularly advantageous in real-time contexts. To assess robustness, we examine scenarios with model misspecification, and verify that the learned decision rules maintain strong clustering accuracy, reflecting the ability of the network to generalize beyond exact model assumptions. An application involving time series of stock returns illustrates the usefulness of the proposed approach for pattern recognition of distinct volatility dynamics. 

Future work can focus on extending the framework to fuzzy clustering, where membership degrees capture partial similarity across temporal patterns, accommodating overlaps common in real-world data like regime-switching series. Hybrid training strategies combining simulation with self-supervised learning on real data may further improve adaptation to empirical distributions without full generative modeling. These directions reinforce the framework’s potential as a flexible, automated tool for uncovering structure in complex time series datasets.
   
\section*{Acknowledgments}
	
The authors thank King Abdullah University of Science and Technology (KAUST) for its support. 

\section*{Code availability statement}

The Python code for running the analysis presented in this paper is available in the GitHub repository: \href{https://github.com/anloor7/PostDoc/tree/main/r_code/amortized_neural_clustering}{https://github.com/anloor7/tests}.

\section*{Data availability statement}

Data used in Section~\ref{sectionapplication} were obtained from the \texttt{R} package \textbf{quantmod}~\citep{quantmod}.

\section*{Disclosure statement}

The authors declare that they have no conflicts of interest relevant to the content of this article.

\bibliography{mybibfile.bib}
	
\end{document}